\documentclass[sn-mathphys-num,iicol]{sn-jnl}

\usepackage{graphicx}%
\usepackage{multirow}%
\usepackage{amsmath,amssymb,amsfonts}%
\usepackage{amsthm}%
\usepackage{mathrsfs}%
\usepackage[title]{appendix}%
\usepackage{xcolor}%
\usepackage{textcomp}%
\usepackage{manyfoot}%
\usepackage{booktabs}%
\usepackage{array}%
\usepackage{algorithm2e}
\usepackage{listings}%
\usepackage{array}
\usepackage{setspace}
\newcolumntype{M}[1]{>{\centering\arraybackslash}m{#1}}%
\usepackage[export]{adjustbox}%
\usepackage[caption=true,listofformat=subsimple,labelformat=simple,font=footnotesize,labelfont=rm,textfont=rm]{subfig}%

\begin{document}

\title[Article Title]{DIFEM: Key-points Interaction based Feature Extraction Module for Violence Recognition in Videos}


\author[1]{\fnm{Himanshu} \sur{Mittal}}\email{himanshumittal.hbti@gmail.com}
\equalcont{These authors contributed equally to this work.}

\author[1]{\fnm{Suvramalya} \sur{Basak}}\email{suvramalya97@gmail.com}
\equalcont{These authors contributed equally to this work.}

\author*[1]{\fnm{Anjali} \sur{Gautam}}\email{anjaligautam@iiita.ac.in}

\affil[1]{\orgdiv{Computer Vision and Biometrics Lab, Department of Information Technology}, \orgname{Indian Institute of Information Technology Allahabad}, \orgaddress{\city{Prayagraj},\state{Uttar Pradesh}, \country{India}}}


\abstract{Violence detection in surveillance videos is a critical task for ensuring public safety. As a result, there is increasing need for efficient and lightweight systems for automatic detection of violent behaviours. In this work, we propose an effective method which leverages human skeleton key-points to capture inherent properties of violence, such as rapid movement  of specific joints and their close proximity. At the heart of our method is our novel Dynamic Interaction Feature Extraction Module (DIFEM) which captures features such as velocity, and joint intersections, effectively capturing the dynamics of violent behavior. With the features extracted by our DIFEM, we use various classification algorithms such as Random Forest, Decision tree, AdaBoost and k-Nearest Neighbor. Our approach has substantially lesser amount of parameter expense than the existing state-of-the-art (SOTA) methods employing deep learning techniques. We perform extensive experiments on three standard violence recognition datasets, showing promising performance in all three datasets. Our proposed method surpasses several SOTA violence recognition methods.}




\maketitle

\section{Introduction}\label{intro}
\vspace{-0.2cm}
Urban violence and crime have become increasingly severe threats to society. To deter such violent activities, the use of surveillance cameras has become widespread in major cities all over the world. As a result, there has been a growing emphasis on the development of automated detection systems for violent activities. Such automated systems not only enhance the capabilities of existing surveillance infrastructure, but also reduce the dependency on human operators, who might miss critical incidents due to fatigue or oversight. In recent years, the application of deep learning and computer vision has revolutionized violence detection, enabling more accurate and efficient identification of violent activities
in crowded and complex environments. Research has focused on extracting meaningful motion information and movement patterns for differentiating between violent and non-violent actions \cite{cheng2021rwf,islam2021efficient,dai2015fudan}. Garcia et al.\cite{garcia2023human} focuses on skeleton features which capture human joint focused motion features. To extract temporal features, \cite{cheng2021rwf,islam2021efficient,garcia2023human} use ConvLSTMs with different modifications. Other works, such as \cite{ullah2023sequential}, use temporal convolution network (TCN) for the same. Several works utilize human skeleton key-points for violence recognition. Matei et al. \cite{matei2025crime} introduce a trajectory-based classification framework, which incorporates skeleton keypoint trajectories. Zhang et al. \cite{zhang2024framework} uses skeleton keypoints to capture motion relations between frames. Specifically, they propose a Weight Selection Module to distinguish between RGB features and skeleton features. Also a Weight Distribution Module is proposed to make the model pay further attention to keyframe information. Tran et al. \cite{tran2024violence} passes the skeleton graph into a Graph Convolution Network (GCN) for violence classification. However, these deep learning-based methods are often very expensive. Further, existing violence datasets are smaller, causing the deep learning models to overfit, thus harming detection accuracy.

Rather than focusing on more complicated and resource intensive deep learning models, we propose a simple yet effective machine learning approach using our relational feature extraction module. It is observed that violent activities can be characterized by rapid and large movement of certain body parts, such as hand, head, feat etc. Hence, we propose the Dynamic Interaction Feature Extractor Module (DIFEM), which extracts interaction information between human pose key-points. The DIFEM calculates the velocity of each joint across consecutive frames, and the amount of overlap of joints of one person with another. This shows that for violence datasets, complicated deep learning models are often not needed. These features are then passed onto statistical classifiers to produce class predictions. Through experiments on three standard datasets, we show that our simple approach achieves comparable or better results compared to existing state-of-art methods which use much larger deep learning models. \par
The main contributions are:
\begin{itemize}
    \item We propose an efficient feature extraction module, named DIFEM, from human pose key-points, leveraging simple heuristics. Change in joint key-point positions across frames gives us the temporal information, while joint overlap between two different human objects give us spatial proximity information.
    \item Machine Learning techniques utilized to classify the DIFEM features, which are significantly less expensive than deep learning methods.
    \item Extensive experiments on three publicly available datasets produce encouraging results, surpassing existing methods. 
\end{itemize}

\section{Literature Review}
\label{literature}
\vspace{-0.2cm}
This section explores various methods and techniques utilized in the existing literature on violence recognition. Relevant features were captured by feature descriptors in traditional methods. Deniz et al. \cite{deniz2014}  detected extreme acceleration patterns by applying random transforms to consecutive video frame power spectrum. 
De Souza et al. \cite{desouza2010b} compared SIFT and STIP, and found that STIP outperformed SIFT on the same datasets.
Xu et al. \cite{xu2014} employed a support vector machine (SVM) with a radial basis function (RBF) kernel for violence detection, and used MoSIFT and STIP for feature extraction. This combination improved the detection of relevant shape and motion patterns.
Zhang et al. \cite{zhang2016} used an SVM classifier with optical flow and a Gaussian model.

\begin{figure*}[!ht]
    \centering
      \includegraphics[width=0.6\textwidth]{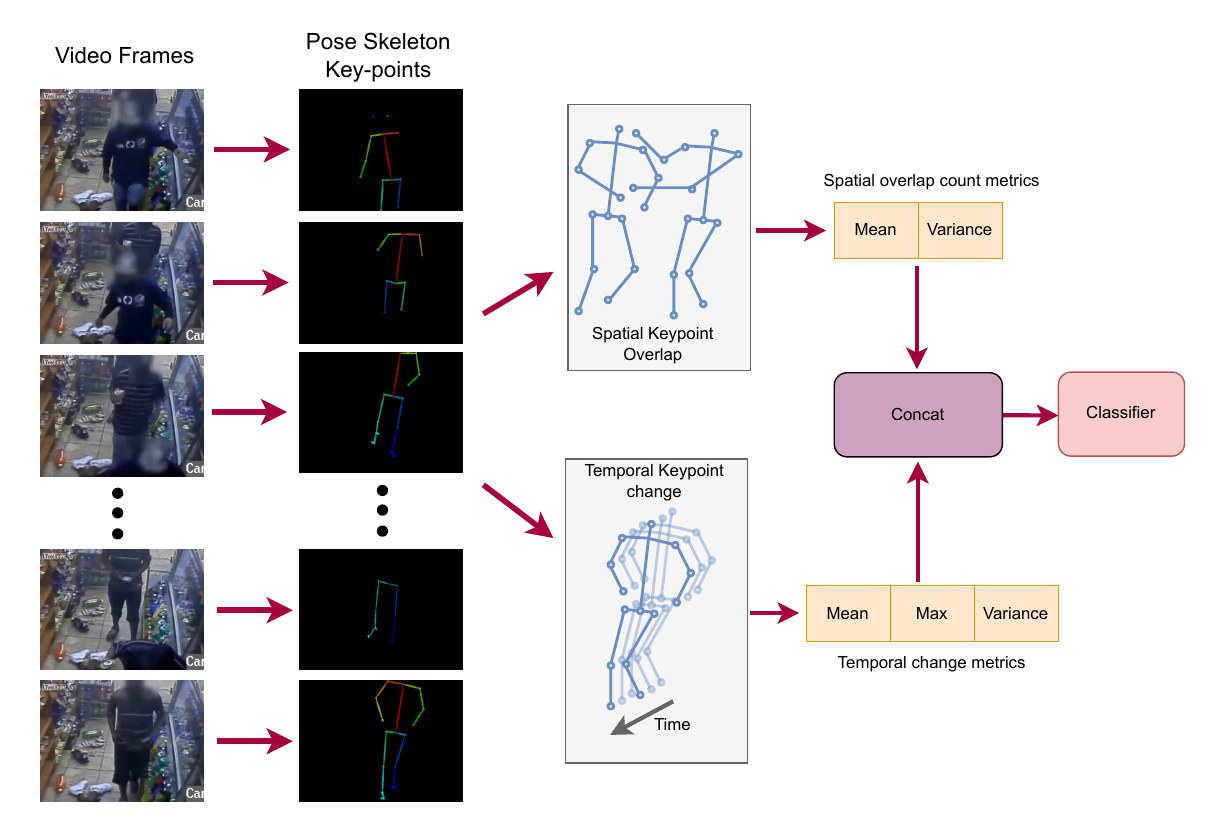}
      \caption{Overview of our proposed approach. The pose key-points of human objects are first extracted using the OpenPose \cite{cao2017realtime} algorithm. From these key-point coordinates, temporal and spatial features are extracted, as explained in Section \ref{DIFEM}. These features, after concatenation, are given to different classifiers.}
      \label{overview}
\end{figure*}

Deep learning techniques have shown remarkable improvements over traditional machine learning methods in violence detection, primarily due to their ability to learn complex spatio-temporal features directly from raw video data. In the existing literature, 3D convolutional networks have been widely used for the detection of violent activities \cite{ding2014violence,tran2015,li2019efficient}. Other works adopted the two-stream approach proposed by \cite{simonyan2014two}, to capture temporal and spatial features for violence recognition \cite{dai2015fudan,cheng2021rwf,islam2021efficient,garcia2023human}. The combination of CNNs and LSTMs for violence detection has shown effectiveness, as evidenced by the work of Sudhakaran and Lanz \cite{sudhakaran2017}, and Soliman et al. \cite{soliman2019b}. Other works \cite{basak2024multi,vaishy2024early} proposed early violence recognition methodology utilizing teacher-student training paradigm.

Other works look to incorporate human skeleton key-points for better violence detection. Jafri et al. \cite{jafri2022skeleton} introduced a skeleton-based deep learning approach to recognize violence in surveillance videos. Matei et al. \cite{matei2025crime} finds the trajectory of each skeleton key-point across frames, and uses this information for a trajectory-based classification framework. Tran et al. \cite{tran2024violence} uses the skeleton key-poins to build a graph. They calculate euclidean distances between the key-points and use this as edge information. This graph is then passed onto a GCN. Our, method differs from these previous works by using simple statistical measures to find task-relevant features, which are provided to machine learning classifiers.

\vspace{-0.2cm}
\section{Proposed Methodology}
\label{methodology}
\vspace{-0.2cm}
In this section, we detail our proposed approach for violence recognition. Unlike existing works \cite{islam2021efficient,garcia2023human,cheng2021rwf,gao2016violence} which use deep learning based models to extract motion and spatial information, we use simple distance metrics to perform the same. The key component of our approach is the Dynamic Interaction Feature Extractor Module (DIFEM), which plays a crucial role in extracting dynamic motion and interaction features. This section is divided into two parts. The details of our DIFEM is explained in Section \ref{DIFEM}. We then utilize the features extracted by the DIFEM to classify video fights using classifiers such as Random Forest, Decision tree, AdaBoost and K-nearest Neighbors.

\vspace{-0.3cm}
\subsection{Dynamic Interaction Feature Extractor Module (DIFEM)}
\label{DIFEM}
\vspace{-0.2cm}
Most violence recognition datasets contains largely human fight videos. Fighting can be characterised by rapid movement of certain human joints. By analyzing how certain body joints move and interact with other joints can provide crucial information which can be helpful for classifying fights. To this end, we propose our Dynamic Interaction Feature Extractor Module. First, we extract skeleton key-point generated by a pre-trained pose estimation algorithm. Skeleton key-points provide us with precise coordinates of human joints, and computing velocities from these key-points allow us to capture the rapid movements effectively. This approach is inspired by previous works that utilized human pose estimation to construct graphs and applied Graph Convolutional Networks (GCNs) for action recognition \cite{su2020human}. However, our method diverges by focusing on extracting and analyzing velocity features directly, offering a simpler yet effective alternative. Additionally, we calculate the joint overlap count, which help us utilize the spatial relationships between the key-points. 
\vspace{-0.3cm}
\subsubsection{Key-points Generation}
\vspace{-0.2cm}
Initially, we will start by extracting key-points from the videos frame wise. Each key-point includes its (x, y) coordinates and confidence level, providing spatial information about detected persons. For each detected person in the frame, we store the key-point information as array. This list of arrays are then saved in the form of a JSON file. Our input data to the DIFEM module consists of JSON files containing key-points detected in every frame of the video. In this work we have used the OpenPose model for key-point extraction. 
\begin{figure}[!ht]
    \centering
      \includegraphics[width=0.26\textwidth]{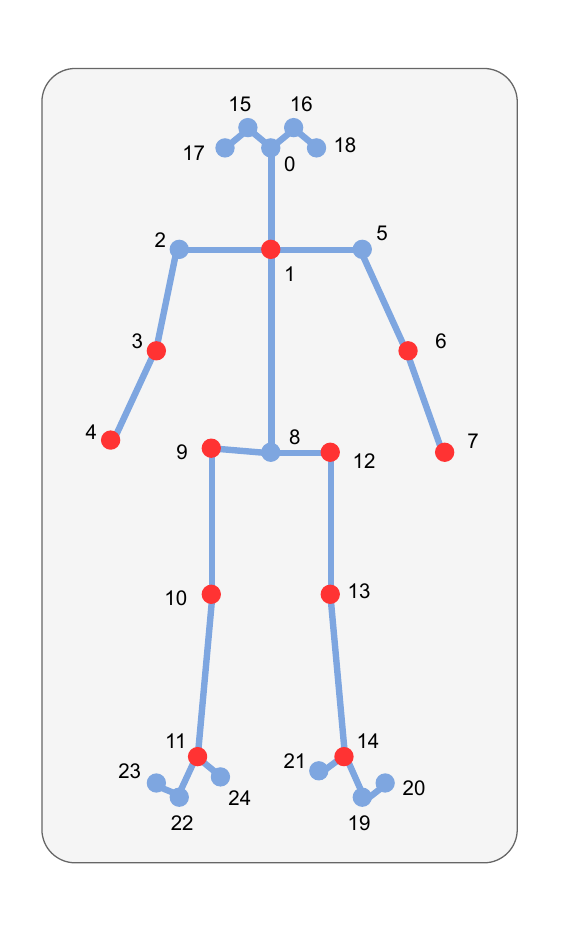}
      \caption{Selected pose key-points, out of 25 OpenPose key-points, we have selected 11 which  are most likely to have significant involvement in physical confrontations. These key-points are marked with a red circle.}
      \label{selected_keypoints}
\end{figure}
\subsubsection{Selection of Key-points}
\vspace{-0.2cm}
From the available key-points, a subset of joints has been strategically chosen to emphasize body parts most pertinent to violence detection. Each joint was selected based on how likely it is to be involved in violent actions. For instance, wrists and elbows tend to exhibit crucial movements indicative of aggressive actions, while hips, ankles, and neck contribute to overall body posture and interaction dynamics. Moreover, specific weights were assigned to each joint to reflect their relative importance in capturing meaningful motion patterns and interactions during violence scenarios. From the 25 key-points extracted by the OpenPose model, we select 11 key-points corresponding to body joints which are most likely to have significant involvement in physical confrontations, ensuring that the most relevant movements are captured. These selected key-points are right wrist, left wrist, right elbow, left elbow, right hip, left hip, right knee, left knee, right ankle, left ankle, and neck. These selected key-points can be seen in Figure \ref{selected_keypoints}. 
\vspace{-0.3cm}
\subsubsection{Feature Calculation}
\vspace{-0.2cm}
Our DIFEM processes the selected skeleton key-points to calculate temporal and spatial dynamics. 
\par

\begin{figure}[t]
    \centering
      \includegraphics[width=0.4\textwidth]{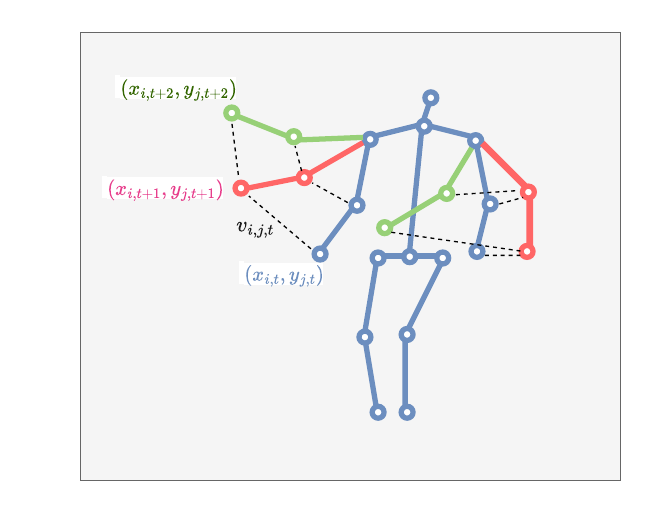}
      \caption{Visualization for calculation of key-point velocity. The skeleton in blue denotes the pose key-points at time $t$, red denotes key-points at time $t+1$, green denotes key-points at time $t+2$. The velocity metric calculates how far a particular key-point $(x_{i,t},y_{j,t})$ at time $t$ moves to at time $t+1$, denoted by $(x_{i,t+1},y_{j,t+1})$.  }
      \label{velocity_diag}
\end{figure}



\RestyleAlgo{ruled}
\IncMargin{-1.5em}
\begin{algorithm}
\caption{Dynamic Interaction Feature Extraction Module (DIFEM)}

\textbf{Input:} Frame-wise key-point coordinates (x, y) for each human person $i$\;
\textbf{Output:} Dataset with extracted features and labels\;
\For{Each video}{
    \For{Each frame}{
        Parse key-points for each detected person $i$\;
        \textbf{Select Joints:} 11 keypoints selected\;
        \textbf{Calculate Velocities} according to equation \ref{velocity_eq}\;
        \textbf{Calculate Joint overlap} according to equation \ref{joint_overlap_eq}\;
    }
    Calculate mean, variance and maximum \par  \textbf{Velocities} across all frames.\;
    Calculate mean and maximum \par \textbf{Joint overlap} across all frames\;
    }
Save the calculated statistical measures as a pickle file for further analysis.
\end{algorithm}

\textbf{Temporal Dynamics} : The DIFEM captures the temporal evolution of joint movements over consecutive frames, discerning rapid and erratic motions indicative of violent behavior by computing the velocities of key-points to identify changes in speed and direction. \par
Let's denote:
\begin{itemize}
    \item $(\mathbf{x}_{i,t}, \mathbf{y}_{i,t})$ as the coordinates of joint $j$ of person $i$ in frame $t$,
    \item $(\mathbf{x'}_{i,t+1}, \mathbf{y'}_{i,t+1})$ as the coordinates of the nearest joint $j$ of person $i$ in frame $t+1$,
    \item $\mathbf{w}_{i,j,t}$ as the weight assigned to joint $j$ of person $i$ in frame $t$.
\end{itemize}

The velocity $\mathbf{v}_{i,j,t}$ for joint $j$ of person $i$ in frame $t$ is calculated as:
\begin{equation}
    \mathbf{v}_{i,j,t} = \sqrt{
    \begin{aligned}
          \mathbf{w}_{i,j,t} \cdot ((\mathbf{x'}_{i,j,t+1} -  \mathbf{x}_{i,j,t})^2 + \\ (\mathbf{y'}_{i,j,t+1} - \mathbf{y}_{i,j,t})^2)
    \end{aligned}
    }
    \label{velocity_eq}
\end{equation}

Fig. 3 shows visualization of how velocity is calculated from key-point movements. We calculate the velocity for all selected joints with their respective weights, and this process is repeated for all consecutive frames in the video. 

Finally after velocities have been calculated for every frame, we take the mean and maximum velocities across all frames of a video. Further, we calculate the amount of variance in the velocity data. The final output is a 1-D vector containing the mean, maximum and variance values. 

\textbf{Spatial Dynamics} : The DIFEM analyzes the spatial
configurations of joint key-points, focusing on close
proximity of joints of two different human object during aggressive
actions.

\begin{figure}[t]
    \centering
      \includegraphics[width=0.5\textwidth]{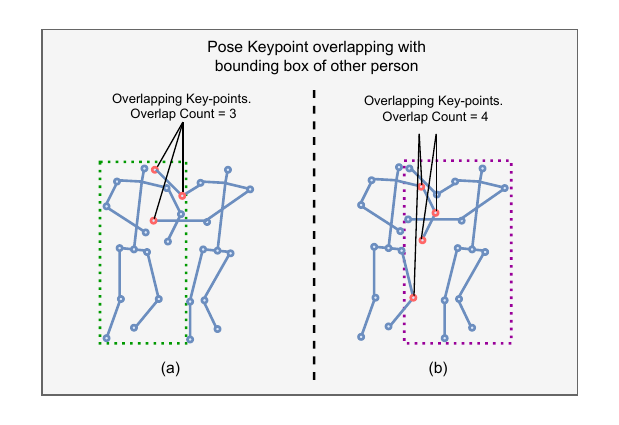}
      \caption{Visualization of key-point overlap measure.(a) The number of overlapping joints (marked in red) is 3; (b) Number of overlapping joints is 4.  }
      \label{overlap_diag}
\end{figure}

The \textbf{Joint overlap measure} calculates the amount of overlap of various joints (such as wrists, elbows, etc.) between different individuals within each frame. It checks if each joint of interest for person 1 falls within the bounding box of person 2, counting the number of such overlaps. This count serves as a measure of the interaction between the two persons on the basis of joint overlaps.

The formula for detecting joint overlap can be expressed as follows.
\vspace{-0.3cm}
\begin{equation}
    \text{JO} = \sum_{j \in \text{joint\_indices}} 
    \begin{cases} 
    1, & \text{if } x_{j}^{(1)} \in [x_{\text{min}}^{(2)}, x_{\text{max}}^{(2)}] \\
      & \text{ and } y_{j}^{(1)} \in [y_{\text{min}}^{(2)}, y_{\text{max}}^{(2)}] \\
    0, & \text{otherwise}
    \end{cases}
    \label{joint_overlap_eq}
\end{equation}

\vspace{-0.2cm}
where:
\begin{itemize}
    \item \( \text{JO is the Joint overlap count} \) which is the number of joint intersections detected in the other person bounding box.
    \item \( j \) iterates over the joint indices in \( \text{joint\_indices} \).
    \item \( (x_{j}^{(1)}, y_{j}^{(1)}) \) are the coordinates of joint \( j \) for person 1.
    \item \( [x_{\text{min}}^{(2)}, x_{\text{max}}^{(2)}] \) and \( [y_{\text{min}}^{(2)}, y_{\text{max}}^{(2)}] \) are the bounding box coordinates of person 2's body.
\end{itemize}

 Similar to temporal dynamics, we calculate the mean number of joint overlaps across all frames in each video, along with the variance of the data-points. The output is a 1-D vector with 2 values.
\vspace{-0.5cm}
\subsection{Violence Classification}
For the task of classifying videos into Violence and Non-Violence classes, we train various machine learning based classifiers on the features extracted by our DIFEM model. We concatenate the features from temporal and spatial dynamics to form a 1-D vector with 5 values. For this work, we try 4 different classification methods; the Nearest Neighbor classifier, AdaBoost, Decision tree and Random Forest.  
\vspace{-0.3cm}
\section{Experimental Details}
\label{exp_details}
\vspace{-0.3cm}
\subsection{Datasets}
We evaluated our method on three standard datasets, the RWF-2000 dataset, the Hockey Fight dataset and Crowd violence dataset.
The RWF-2000 dataset \cite{cheng2021rwf} is a large-
scale video dataset comprising of 2,000 videos, with 800 violence and 800 non-violence videos. 
The Hockey fight \cite{nievas2011} contains 1000 videos of actions from hockey match, with 500 Fight and 500 Non-Fight videos.
The crowd violence dataset \cite{hassner2012violent} contains 246 real world video clips, out of which 123 videos are of normal behaviour and 123 videos contain violence behaviour. 

\vspace{-0.3cm}
\subsection{Evaluation Metrics}
Following existing studies \cite{cheng2021rwf,garcia2023human,islam2021efficient}, we use the accuracy measure to evaluate our model. The accuracy metric measures what percentage of all the predicted class outputs are correct. We further reported the precision, recall and f1-score of our method on the two datasets. Precision metric measures what percentage of the positive output predictions are correct. Recall measures what percentage of the total number of actual positive classes were predicted correctly as positive. F1-score is a combination of the precision and recall, measures how well our method can balance the trade-off between the two.
\vspace{-0.3cm}
\subsection{Implementation Details}
We have utilized  OpenPose \cite{cao2017realtime} model for pose estimation, as it is able to detect multi-person poses. Additionally, the computational time for the OpenPose is not dependent on the person count in the video, which makes it beneficial for real-time pose estimations. When selecting important keypoints, we set weights as follows; [``right wrist", ``left wrist", ``right elbow", ``left elbow", ``right hip", ``left hip", ``right knee", ``left knee", ``right ankle", ``left ankle", ``neck"] = [1, 1, 0.8, 0.8, 1, 1, 1, 1, 1, 1, 1]. For the implementation of our machine learning models, we use the scikit-learn library. For random forest, we build 100 decision trees with Gini-impurity measure. For K-nearest neighbors, we take $k=5$. We set number of estimators to 100 for AdaBoost classifier. For decision tree classifier, Gini-impurity measure is used with no max depth. Other parameters are set to default for all classifiers.

\begin{table}[h!]\vspace{-0.5cm}
\caption{Comparison with state-of-the-art on the RWF-2000, Hockey-Fight and Crowd datasets.}\label{comp_all}%
\centering
\begin{tabular}{@{}llll@{}}
\toprule
Method & RWF & Hockey & Crowd  \\
\midrule
ViF \cite{hassner2012violent} & - & 82.90 & 85.00  \\
ViF + OViF \cite{gao2016violence} & - & 87.50 & 88.00 \\
HOG+HIK \cite{bermejo2011violence} & - & 91.70 & - \\
MoWLD+BoW \cite{zhang2017mowld} & - & 91.90 & 82.56 \\
LHOG+LOF \cite{zhou2018violence} & - & 95.10 & 94.31 \\
\hline
TSN \cite{wang2016temporal} & 81.5 \footnotemark[1] & 91.5 \footnotemark[1] & 81.5 \footnotemark[1] \\
I3D \cite{carreira2017quo} & 83.4 \footnotemark[1] & 93.4 \footnotemark[1] & 83.4 \footnotemark[1]  \\
3D-ResNet101 \cite{hara2018can} & 82.6 \footnotemark[1] & - & -   \\
3D CNN \cite{ding2014violence} & - & 82.6 \footnotemark[1] & - \\
Cheng et al. \cite{cheng2021rwf} & 87.3 & 98.0 & 88.8  \\
DGCNN \cite{wang2019dynamic} & 80.6 \footnotemark[1] & 90.2 \footnotemark[1] & 87.4 \footnotemark[1] \\
SPIL \cite{su2020human} & 89.3 & 96.8 & 94.5  \\
Islam et al. \cite{islam2021efficient} & 89.75 & 99.0 & - \\
Garcia-Cobo et al. \cite{garcia2023human} & 90.25 & 94.50 & 94.30  \\
Hachiuma et al. \cite{hachiuma2023unified} &  93.4 & 99.5 & 94.7\\
Ullah et al. \cite{ullah2023sequential} & 91.15 & - & 96.0\\
Zhang at al. \cite{zhang2024framework} & 93.3 & 99.4 & - \\
Tran et al. \cite{tran2024violence} & 65.68 & 71.39 & - \\
\midrule
Ours (DIFEM+Random Forest) & 96.50 & 98.80 & 96.33 \\
Ours (DIFEM+Decision Tree) & 91.75 & 97.10 & 93.47  \\
Ours (DIFEM+AdaBoost) & 95.50 & 98.50 & 94.30\\
Ours (DIFEM+Nearest Neighbor) & 91.75 & 98.70 & 94.30\\
\hline
\end{tabular}
\footnotetext[1]{Results as reported in \cite{su2020human}}
\end{table}

\begin{table*}[h!]
\tiny
\caption{Class-wise Precision, Recall and F1-scores for our method on RWF-2000, Hockey-Fight and Crowd datasets.}\label{class_report}%
\centering
\begin{tabular}
{M{1.3cm}|M{1.2cm}|M{1.2cm}|M{0.9cm}|M{0.9cm}|M{1.2cm}|M{0.9cm}|M{0.9cm}|M{1.2cm}|M{0.9cm}|M{0.9cm}}
\toprule
& & \multicolumn{3}{c|}{RWF-2000} & \multicolumn{3}{c|}{Hockey-Fight} & \multicolumn{3}{c}{Crowd}\\
\cmidrule{3-11} 
Method & Class & Precision & Recall & F1 & Precision & Recall & F1 & Precision & Recall & F1\\
\midrule
\vspace{2mm}
\multirow{2}{4.5em}{K-Nearest Neighbor} & NonFight & 92.0 & 91.0 & 92.0 & 99.6 & 97.8 & 98.6 & 95.2 & 93.6 & 94.4 \\
& Fight & 91.0 & 93.0 & 92.0 & 97.8 & 99.6 & 98.8 & 93.6 & 95.2 & 94.4  \\
\hline
\vspace{2mm}
\multirow{2}{4.5em}{Decision Tree} & NonFight & 92.0 & 92.0 & 92.0 & 96.6 & 97.6 & 97.2 & 93.0 & 94.4 & 93.6 \\
& Fight & 92.0 & 92.0 & 92.0 & 97.6 & 96.6 & 97.0 & 94.4 & 92.8 & 93.6 \\
\hline
\vspace{2mm}
\multirow{2}{4.5em}{AdaBoost} & NonFight & 98.0 & 93.0 & 95.0 & 98.8 & 98.2 & 98.6 & 95.0 & 93.0 & 93.8 \\
& Fight & 93.0 & 98.0 & 96.0 & 98.2 & 98.8 & 98.4 & 93.2 & 95.2 & 94.0 \\
\hline
\vspace{2mm}
\multirow{2}{4.5em}{Random Forest} & NonFight & 99.0 & 94.0 & 96.0 & 98.2 & 99.4 & 99.0 & 95.4 & 97.6 & 96.4 \\
& Fight & 94.0 & 99.0 & 97.0 & 99.4 & 98.2 & 98.6 & 97.6 & 95.2 & 96.4\\
\hline
\end{tabular}
\end{table*}

\section{Results and Discussions}
\label{results}
\vspace{-0.2cm}
\subsection{Comparative analysis}
\begin{figure*}[!ht]
    \centering
      \includegraphics[width=0.9\textwidth]{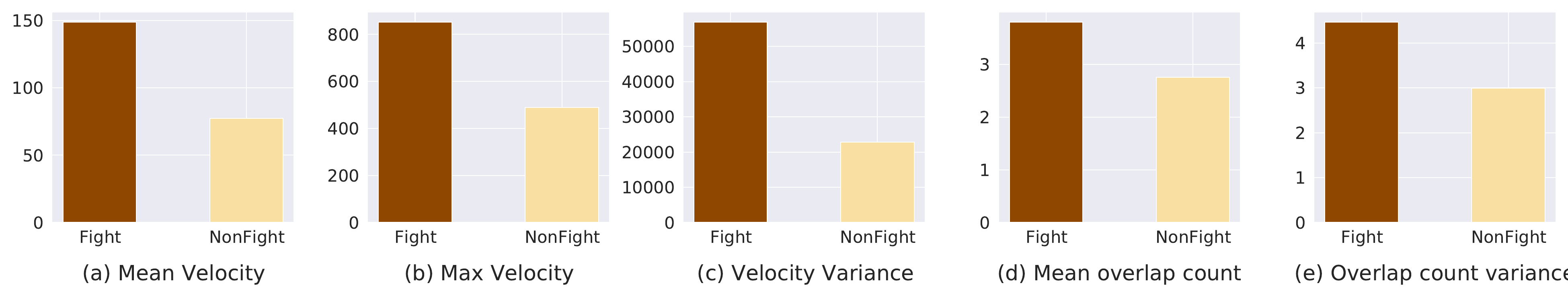}
      \caption{Velocity and overlap measures averaged over all videos in test set of RWF-2000 dataset. ``Fight" videos, on average, has higher velocity and joint overlap count than ``Non-Fight" videos. }
      \label{DIFEM_feature_values}
\end{figure*}

We compare the results of our method with those of existing works. The comparative analysis is given in Table \ref{comp_all}. The first five works in the table are traditional approaches, whereas the rest are deep learning-based. Our approach using random forest achieves the highest performance. We have also shown the results of our DIFEM feature extractor with other machine learning model, namely decision tree, AdaBoost and Nearest Neighbor classifiers. 
For Hockey-Fight and Crowd datasets we report 5-fold cross validation average of our method. The class-wise precision, recall and F1-score for our method on all three datasets are listed in Table \ref{class_report}. 
Further, we show the average feature values for all videos present in the test set of RWF-2000 in Figure \ref{DIFEM_feature_values}. These results demonstrate the effectiveness of our approach on multiple datasets, achieving state-of-the-art in various standard datasets. These results validate our initial assumption that for violence recognition datasets, the amount of movement of joints acts as the most important feature. As seen from Figure \ref{DIFEM_feature_values}, videos of ``Violence'' class contains more large and rapid joint movements than videos in ``Non-Violence'' class. Hence, with proper estimation of the amount of movement and overlap of human joints, the requirement of more computationally expensive deep learning models diminishes. In contrast to other deep learning based methods, our method utilizes simple heuristics to extract spatio-temporal features, followed by traditional machine learning algorithms for classification, making it more computationally economical. Even similar works using skeleton keypoints \cite{tran2024violence,zhang2024framework} utilize larger neural networks to process skeleton joint movements. 

To show the efficiency of our approach, we calculate the number of tree nodes for Random forest, decision tree, AdaBoost classifiers, which are 13202, 161, 300, respectively. Nearest neighbor is a non-parametric classifier. Additionally, the average test time of each video in test set for Random forest, decision tree, AdaBoost and Nearest neighbor are 0.0333 milli-second (ms), 0.0232 ms, 0.0532 ms and 0.0533 ms,respectively.

\begin{table}[h]
\caption{Ablation studies of our method.}\label{ablation}%
\centering
\begin{tabular}{M{1.2cm}M{1.2cm}|M{1.0cm}|M{1.0cm}|M{1.2cm}}
\toprule
Velocity Measure & Joint overlap measure & RWF & Hockey Fight & Crowd Violence\\
\midrule
\centering\checkmark & & 91.0 & 92.2 & 89.04 \\
& \centering\checkmark & 86.0 & 82.6 & 73.16\\
\centering\checkmark & \centering\checkmark  & 96.5 & 98.80 & 96.33\\
\hline
\end{tabular}
\end{table}

\subsection{Ablation Studies}
We perform ablation study to demonstrate the efficacy of our DIFEM feature extractor on the three datasets. The results can be seen in Table \ref{ablation}. Using only the key-point overlap measure, we are only able to achieve spatial relations between human objects. However, joints may also overlap in case of human occlusion. Hence, this may lead to erroneous predictions. Further, using only joint overlap measure, we are unable to utilize temporal information. These limitations are especially evident on Crowd dataset. This is because of the crowded scenarios of the dataset, which leads large number of joint overlaps that are not related to violence. In such situations, using the velocity metric helps differential the classes. The velocity metric adds temporal information by finding how rapid joints are moving in consecutive frames. Using a combination of the two provides us with the best results for all three datasets.   

\begin{table*}[h!]
\scriptsize
\centering
\caption{Qualitative results of our method on the RWF-2000 dataset.} \label{Qualitative}
\begin{tabular}{M{8.8cm} | M{1.8cm} | M{1.8cm}}
\hline
\textbf{Video Frames} & \textbf{Ground Truth} & \textbf{Predicted Label} \\
\hline
\includegraphics[width=0.5\textwidth,valign=c]{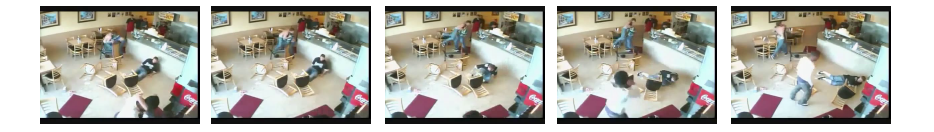}  & Violence & Violence \\
\includegraphics[width=0.5\textwidth,valign=c]{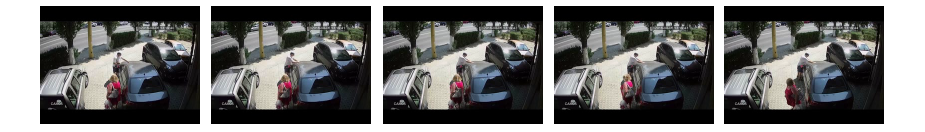}  & Non-Violence & Non-Violence \\
\includegraphics[width=0.5\textwidth,valign=c]{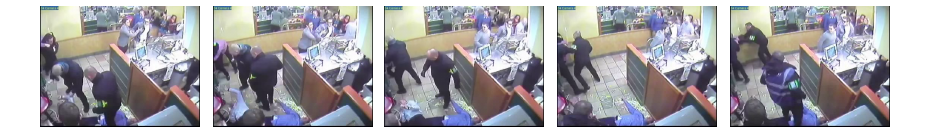}  & Violence & Non-Violence \\
\includegraphics[width=0.5\textwidth,valign=c]{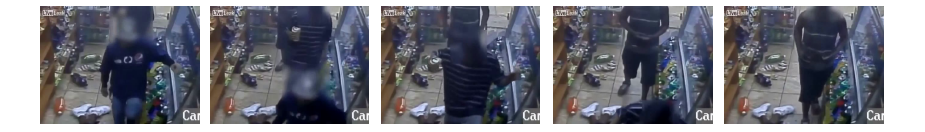}  & Non-Violence & Violence \\
\hline

\end{tabular}
\end{table*}

\subsection{Qualitative analysis}
\vspace{-0.2cm}
The qualitative results of our proposed violence recognition approach is shown in Table \ref{Qualitative}. The first three rows show examples of videos which have been correctly classified by our approach. The last three rows show incorrect predictions of our method. The video presented in row 4 of Table \ref{Qualitative} depicts a violent action. However, the video has less movement, as a result, our DIFEM method gives low velocity scores. This results in the classifier wrongly predicting the video to be of ``Non-violence" class. Similarly, the videos depicted in rows 4 and 5 are from the ``Non-Violence" class. Although these video have low average joint velocities and overlaps, due to large walking strides, DIFEM calculates large maximum joint velocities.     

\section{Conclusion}
\label{conclusion}
\vspace{-0.3cm}
In this paper, a novel feature extraction method is presented for the task of violence recognition. Our proposed DIFEM utilizes pose key-point coordinates from OpenPose algorithm to calculate the change in joint positions across consecutive frames. Further, we have checked the number of times important joints of a person overlap with the bounding box of another person, giving us spatial joint interaction information. Using these features, we employ classifiers such as Random Forest, AdaBoost, k-Nearest Neighbor and Decision Tree to classify the videos. Experimental results show promising results, surpassing existing methods that employ much more complex deep learning models. The results indicate our approach as an effective alternative to much more resource intensive deep learning approaches. \par
The main drawback of our approach is its dependence on the OpenPose algorithm. The accuracy of detection of skeleton key-points effects the down-stream task of feature extract by our DIFEM. Furthermore, similar to existing works using skeleton key-points, the OpenPose algorithm adds a large number of parameters to the overall model. As future work, we plan to work on more parameter efficient skeleton key-point algorithms.

\section*{Declarations}

\textbf{Funding} There are no funding to report. \\
\textbf{Conflict of interest} The authors declare no conflict of interest.\\
\textbf{Ethics approval and consent to participate} Not Applicable.\\
\textbf{Consent for publication} Results/data/figures in this manuscript have
not been published elsewhere, nor are they under consideration by another publisher.\\
\textbf{Data availability} Data is publicly available.\\


\bibliography{references}


\end{document}